\documentclass[numbers]{article}

    \usepackage[preprint]{neurips_2024}

\usepackage[utf8]{inputenc} 
\usepackage[T1]{fontenc}    
\usepackage{hyperref}       
\hypersetup{colorlinks, breaklinks, citecolor=[rgb]{0, 0.44, 0.737},
anchorcolor=[rgb]{1, 0.44, 0.737}, linkcolor=[rgb]{0, 0.44, 0.737},
urlcolor=[rgb]{0, 0.44, 0.737} 
}
\usepackage{url}            
\usepackage{booktabs}       
\usepackage{amsfonts}       
\usepackage{nicefrac}       
\usepackage{microtype}      
\usepackage[dvipsnames]{xcolor}         
\usepackage{subfig}
\usepackage{amsmath, amssymb}
\usepackage{multirow}
\usepackage{tablefootnote}
\usepackage[font=small,labelfont=bf,tableposition=top]{caption}
\DeclareCaptionLabelFormat{andtable}{#1~#2  \&  \tablename~\thetable}
\usepackage{enumitem}
\usepackage{makecell}
\usepackage{wrapfig}

\usepackage{tcolorbox}
\definecolor{lightblue}{HTML}{18282e}
\definecolor{lighterblue}{HTML}{f2fafd}  
\newtcolorbox{abox}{colback=lighterblue,colframe=lightblue}

\newcommand{\ie}{\textit{i.e.}}
\newcommand{\eg}{\textit{e.g.}}

\newcommand{\mtx}[1]{\mathbf{#1}}
\renewcommand{\vec}[1]{\mathbf{#1}}
\DeclareMathOperator{\reals}{\mathbb{R}}

\title{ReSpike: Residual Frames-based Hybrid Spiking Neural Networks for Efficient Action Recognition}

\author{%
  Shiting Xiao\\
  Yale University\\
  {\small \texttt{ginny.xiao@yale.edu}}
  \And
  Yuhang Li\\
  Yale University\\
  {\small \texttt{yuhang.li@yale.edu}}
\And
Youngeun Kim \\
Yale University \\
  {\small \texttt{youngeun.kim@yale.edu}}\\
\And
Donghyun Lee\\
Yale University \\
 {\small \texttt{donghyun.lee@yale.edu}}\\
\And
Priyadarshini Panda \\
Yale University \\
{\small \texttt{priya.panda@yale.edu}}
}

\begin{document}

\maketitle

\begin{abstract}
Spiking Neural Networks (SNNs) have emerged as a compelling, energy-efficient alternative to traditional Artificial Neural Networks (ANNs) for static image tasks such as image classification and segmentation.
However, in the more complex video classification domain, SNN-based methods fall considerably short of ANN-based benchmarks due to the challenges in processing dense frame sequences.
To bridge this gap, we propose \textbf{ReSpike}, a hybrid framework that synergizes the strengths of ANNs and SNNs to tackle action recognition tasks with high accuracy and low energy cost. By decomposing film clips into spatial and temporal components, i.e., RGB image {\bf Key Frames} and event-like {\bf Residual Frames}, ReSpike leverages ANN for learning spatial information and SNN for learning temporal information.
In addition, we propose a multi-scale cross-attention mechanism for effective feature fusion.
Compared to state-of-the-art SNN
baselines, our ReSpike hybrid architecture demonstrates significant performance improvements (e.g., $>$\textbf{30}\% absolute accuracy improvement on HMDB-51, UCF-101, and Kinetics-400). 
Furthermore, ReSpike achieves comparable performance with prior ANN approaches while bringing better accuracy-energy tradeoff. Code is shared at \url{https://github.com/GinnyXiao/ReSpike}.
\end{abstract}

\section{Introduction}
\label{sec:intro}

Artificial Neural Networks (ANNs) have demonstrated remarkable performance across a range of computer vision tasks, including object classification~\cite{he2016deep, krizhevsky2012imagenet, szegedy2015going}, detection~\cite{ren2015faster}, and segmentation~\cite{he2017mask}, albeit at significant computing costs, e.g. state-of-the-art (SOTA) Vision Transformers~\cite{dosovitskiy2020image, liu2021swin} require 60-200 GFLOPs of computation per inference~\cite{xu2023devit}. The growing importance of low-power or battery-constrained devices in various real-world applications such as medical robots~\cite{bing2018survey}, self-driving cars~\cite{ zhou2020deep}, and drones~\cite{stagsted2020towards} necessitates energy-efficient neural network architectures. 
Spiking Neural Networks (SNNs) offer a promising alternative~\cite{roy2019towards, christensen20222022, wu2018spatio, wu2022brain, fang2021deep, kim2021visual, kim2023exploring}. Drawing inspiration from the biological neuron, SNNs operate by processing visual information through discrete spikes distributed across multiple timesteps. This distinctive event-driven behavior not only aligns with the intricacies of neural processing but also results in a substantial reduction in energy consumption, especially when implemented on low-power neuromorphic chips~\cite{akopyan2015truenorth, davies2018loihi, furber2014spinnaker}. 

\begin{figure}[t]
\subfloat[ImageNet]{\includegraphics[width=0.325\textwidth]{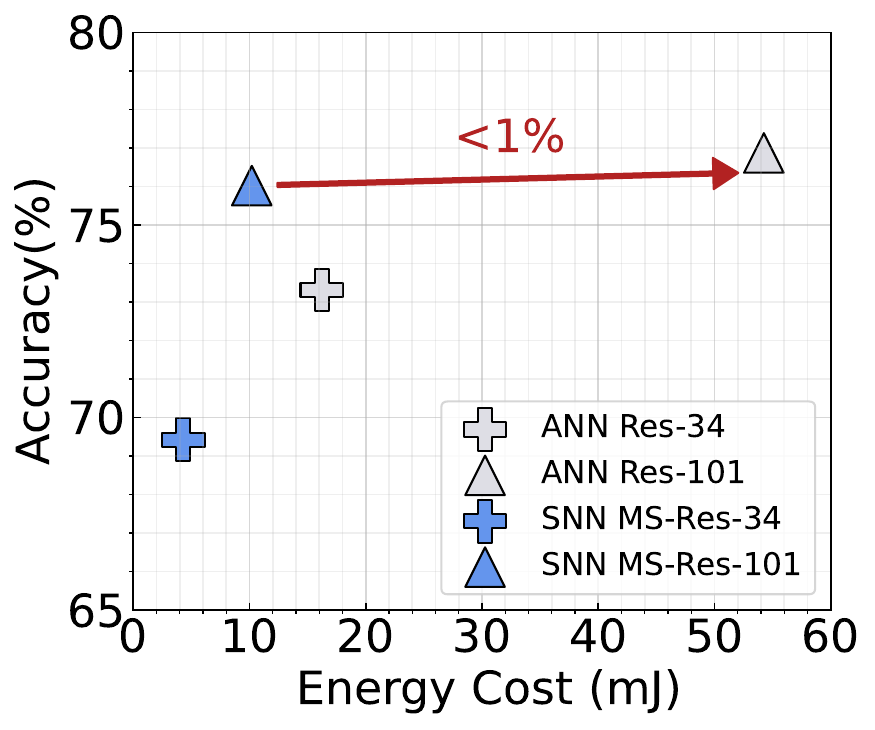}}\hfill 
\subfloat[HMDB-51]{\includegraphics[width=0.33\textwidth]{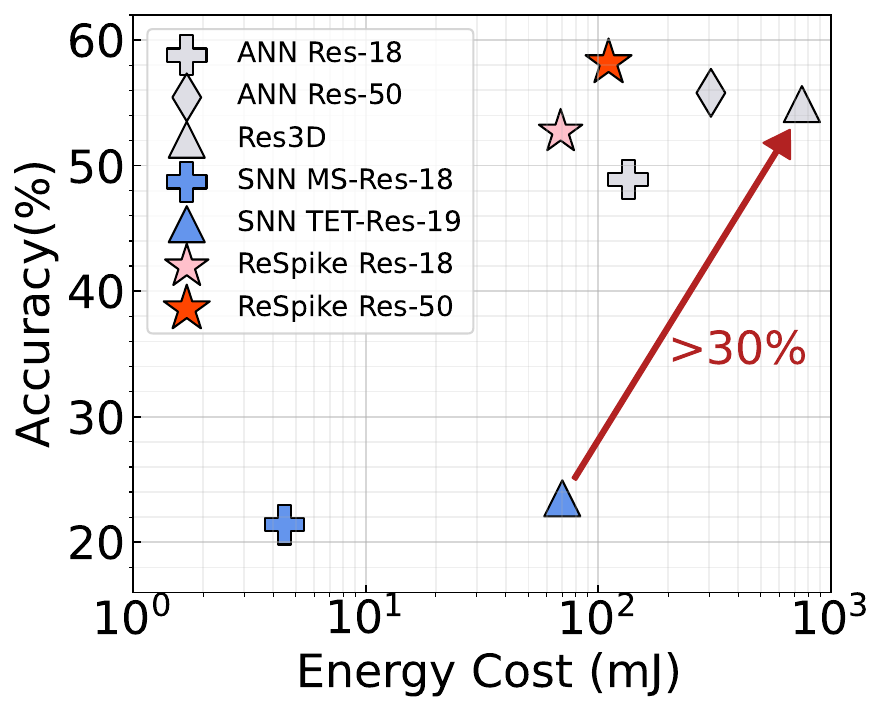}}\hfill
\subfloat[UCF-101]{\includegraphics[width=0.33\textwidth]{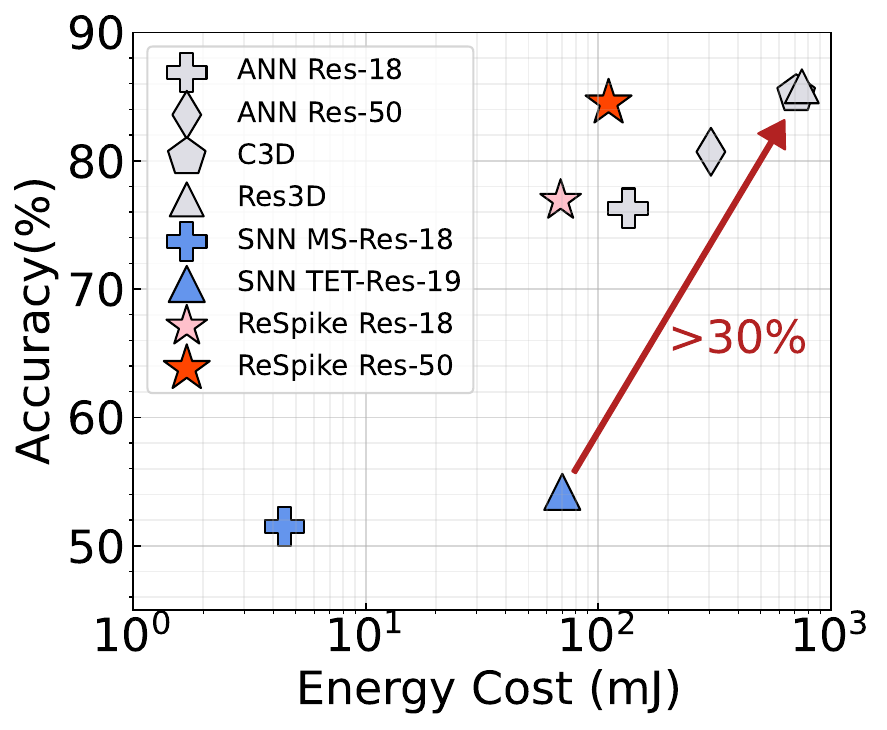}}
\caption{Accuracy and energy consumption for ANN and SNN models on (a) ImageNet, (b) HMDB-51, and (c) UCF-101 datasets. While SOTA SNN models demonstrate comparable accuracy and reduced cost on static ImageNet, they still face a significant performance shortfall compared to ANNs on dynamic action datasets.}
\label{fig:ann-snn-comparisn}
\vspace{-1em}
\end{figure}

In the existing body of literature, SNNs have shown efficacy in static image recognition tasks~\cite{roy2019towards, hu2024advancing, deng2022temporal}, e.g. \cite{hu2024advancing} reports $<1\%$ accuracy gap between the spiking MS-ResNet-101 and the ANN ResNet-101 architectures on the ImageNet~\cite{deng2009imagenet} dataset (Fig. \ref{fig:ann-snn-comparisn}a). However, in a more challenging domain of action recognition, there lacks an SNN-based benchmark that achieves comparable performance against their ANN counterpart. Our preliminary investigations into extending SOTA SNN models~\cite{hu2024advancing, deng2022temporal} for frame-by-frame processing reveal a significant {\bf accuracy gap of $>$30\%} against ANN-based methods~\cite{he2016deep, tran2015learning, tran2017convnet}.
In Fig. \ref{fig:ann-snn-comparisn}b, c, we show an accuracy-energy tradeoff between SNNs and ANNs on two human action benchmarks, HMDB-51~\cite{kuehne2011hmdb} and UCF-101~\cite{soomro2012ucf101}. A similar performance gap is also reported in prior works~\cite{samadzadeh2023convolutional, wang2023integrating}, underscoring the inherent difficulties SNNs face when processing dense sequences of images. Using spikes as signals, SNNs depend heavily on prolonged data exposure to generate predictions on par with ANNs, particularly struggling with the extraction of nuanced information from visually redundant images.

To address this challenge, we propose {\bf ReSpike}, a hybrid neural network framework that leverages the strengths of both ANNs and SNNs to perform high-accuracy and low-energy action recognition.
As SNNs excel in handling sparse and time-series data like event stream~\cite{zhang2022spiking, lee2020spike, zhu2022event} and ANNs prove adept at extracting patterns from dense and spatial data like RGB images, we use a data representation that aligns with their distinctive computing paradigms. Specifically, we decompose a conventional film clip into a spatial component and a temporal component: the spatial part, the same as the original RGB images, carries appearance information about scenes and objects; and the temporal part, in the form of sparse illumination changes across clip, convey the movements of the objects. We call the RGB images {\bf Key Frames} and event-like images {\bf Residual Frames}
. We design ReSpike to integrate an ANN branch for learning spatial features from key frames and an SNN branch for capturing temporal dynamics from residual frames. In addition, we propose a multi-scale cross-attention mechanism for information fusion between the ANN features and SNN features, building correlations between pixels across key and residual at all scales. By strategically combining ANNs and SNNs, our hybrid model achieves a better accuracy-energy tradeoff.

In summary, our contributions center on the challenges mentioned previously:
\begin{enumerate}[leftmargin=*, nosep]
\item We propose a novel framework {\bf ReSpike} featuring a hybrid SNN-ANN feature extractor. 
We introduce a cross-attention fusion module that adeptly integrates ANN's spatial features and SNN's temporal features, enabling ReSpike to capture complementary insights.
\item Compared to prior SOTA SNN works, ReSpike achieves {\bf 31}\%, {\bf 34}\%, and {\bf 32}\% accuracy improvements on three challenging datasets, UCF-101~\cite{soomro2012ucf101}, HMDB-51~\cite{kuehne2011hmdb}, and Kinetics-400~\cite{kay2017kinetics}, respectively. ReSpike marks a pioneering breakthrough as the first SNN method to scale to the large-scale Kinetics-400 benchmark~\cite{kay2017kinetics}, achieving \textbf{70.1}\% accuracy. ReSpike also achieves superior or comparable performance with prior ANN approaches with up to {\bf 6.8$\times$} reduction in energy consumption. In addition, we conduct in-depth ablation studies and plot the visualization results of the attention maps,  providing insightful analysis on hybrid modeling.
\end{enumerate}

\section{Related Works}
\label{sec:related_works}

\textbf{Spiking Neural Networks.} 
SNNs draw inspiration from biological brains, using discrete events or spikes for information processing. This approach, particularly efficient for temporal data, is exemplified in the Leaky Integrate-and-Fire (LIF) model~\cite{cao2015spiking, roy2019towards, wu2019direct}. 
Recent innovations like surrogate gradient learning~\cite{neftci2019surrogate, lee2016training, yang2023snib}, spatial-temporal backpropagation \cite{rathi2020enabling, zheng2021going, kim2021revisiting, rathi2020diet, fang2021incorporating, zhang2021rectified, deng2022temporal, guo2023membrane} and various deep learning architectures including Spiking ResNet~\cite{hu2021spiking, fang2021deep, hu2024advancing, kim2024rethinking} and Spiking Transformers~\cite{zhou2022spikformer, yao2024spike, zhou2023spikingformer} have been developed to directly train larger and deeper SNNs, which have paved the way for SNNs’ application in computer vision tasks, including image classification~\cite{tavanaei2019deep, hu2024advancing, fang2021deep}, segmentation~\cite{kim2022beyond}, object detection~\cite{su2023deep}, and event-based sensing~\cite{yao2021temporal, wang2024eas, zhu2024tcja, zou2023event, zhu2022event}. 

However, few studies have developed SNN-based architectures for dynamic scene classification, revealing a gap in research that mainly focuses on relatively simple static image tasks. 
This limitation results from either constrained applicability or challenges in achieving optimal performance. 
Notably, works like \cite{panda2018learning, el2023spiking, el2023s3tc} evaluated their models on limited datasets such as Weizmann~\cite{blank2005Weizmann}, UCF sports~\cite{rodriguez2008ucfsports}, and KTH~\cite{schuldt2004kth},  featuring only 9, 9, and 6 classes 
respectively. 
Contrarily, we test our model on comprehensive datasets HMDB-51, UCF-101, and Kinetics-400 which comprise 51, 101, and 400 classes,
respectively. 
\cite{samadzadeh2023convolutional, wang2023integrating} are the only prior attempts to develop directly trained SNNs for large datasets. However, they struggle with suboptimal performance, 
primarily due to sequential processing of dense RGB images each for one timestep, leading to redundant computations on visually similar background regions and insufficient processing on subtle motion details. 


\textbf{Hybrid neural networks.}
There is a growing trend to design hybrid neural networks (HNNs) by combining SNNs and ANNs to leverage the strengths of both. Recent studies have shown HNNs' potential in flow estimation~\cite{lee2022fusion, lee2020spike}, tracking~\cite{zhang2022spiking, zhao2022framework}, and human pose estimation~\cite{aydin2023hybrid}, showcasing their ability to achieve superior performance in terms of speed, accuracy, and energy consumption compared to their single-model counterparts.
However, integrating SNNs' temporal modeling capabilities in action recognition is still underexplored. 
 Drawing inspiration from the recent successes of Transformers~\cite{vaswani2017attention} in the hybrid learning context, this paper seeks to bridge this gap, building the inter-modal correlation between SNNs and ANNs through cross-attention mechanism~\cite{jaegle2021perceiver, jaegle2021perceiverio, rombach2022high}.

\section{SNN Preliminaries}
\label{sec:preliminaries}

{\bf Leaky Integrate-and-Fire neuron. }
The primary difference between an SNN and an ANN originates from their basic computing unit, \ie the neuron.
In traditional ANNs, the neuron model is often based on a simple nonlinear activation function, such as the rectified linear unit (ReLU) function \cite{krizhevsky2012imagenet}, given by $\vec{y} = \max (0, \mtx{W}\vec{x})$. 
The behavior of an SNN neuron is often modeled using the leaky integrate-and-fire (LIF) model \cite{cao2015spiking}. Formally, at timestep $t$, when an input signal $\vec{i}^{(t+1)} = \mtx{W}\vec{x}^{(t+1)}$ from the pre-synaptic neurons is given to the LIF neuron, the dynamics of its membrane potential $\vec{u}^{(t)}$ can be described by: 
\begin{align}
    \vec{u}^{(t+1), \text { pre }} &= \vec{i}^{(t+1)} + \tau \vec{u}^{(t)}\\
    \vec{y}^{(t+1)}&=\left\{\begin{array}{ll}
    1 & \quad \text { if } \vec{u}^{(t+1), \text { pre }}>v_{t h} \\
    0 & \quad \text { otherwise }
    \end{array}\right.\\
    \vec{u}^{(t+1)}&=\vec{u}^{(t+1), \text { pre }} \left(1-\vec{y}^{(t+1)}\right)
\end{align}
Here, $ \vec{u}^{(t+1), \text { pre}}$ is the pre-synaptic membrane potential that consists of the weighted spike signals from the pre-synaptic neurons and the decayed membrane potential from previous timesteps, where $\tau$ is a time constant for the decay. When the membrane potential exceeds a firing threshold $v_{t h}$, the LIF neuron fires a spike $\left(\vec{y}^{(t+1)} = 1\right)$; otherwise, it stays inactive $\left(\vec{y}^{(t+1)} = 0\right)$. After firing, the membrane potentials $\vec{u}^{(t+1)}$ reset to 0 and the spike outputs $\vec{y}^{(t+1)}$ are propagated to the next layer.

\textbf{Deep SNNs for temporal tasks. }
SNNs are naturally suitable for temporal tasks such as event stream processing and signal processing~\cite{li2023efficient}, as they have a temporal dimension that allows for spatio-temporal extraction of features.  Given a sequence of temporal data $\{ \vec{x}_t\}_{t=0}^T$ from timestep $t=0$ to $t=T$, SNN sequentially processes each of them in time order. Until a spike occurs, the membrane potentials of all SNN neurons persist in their previous status instead of resetting to the initial potential status, which allows for capturing temporal cues. As the data propagates through the network, temporal integration of input signals occurs inside the LIF neurons, capturing the temporal dependencies between data points. 


\begin{figure}[t]
    \centering
    \includegraphics[width=\textwidth]{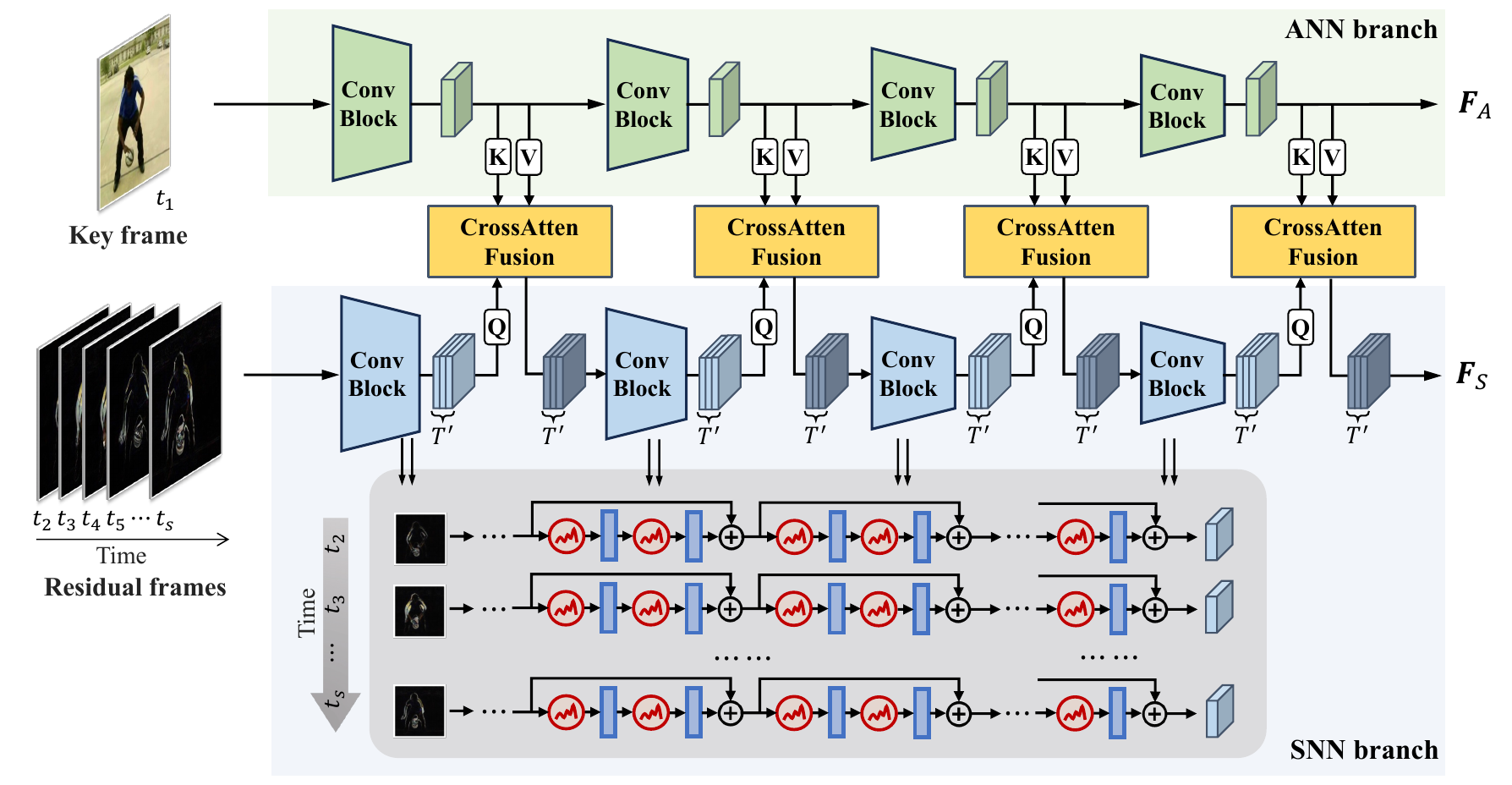}
    \caption{{\bf ReSpike architecture}. We use RGB key frames and event-like residual frames as inputs to our ReSpike model, which consists of an SNN branch, an ANN branch, and four cross-attention modules. 
    The ANN branch output $\vec{F}_A$ and SNN branch output $\vec{F}_S$ will be combined in the classification head for final prediction. }
    \label{fig:network-arch}
    \vspace{-1em}
\end{figure}

\section{ReSpike Methodology}\label{sec:method}
We propose \textbf{ReSpike}, a hybrid architecture that performs both spatial feature extraction and temporal dynamics modeling for action recognition in an efficient manner.
As depicted in Fig. \ref{fig:network-arch}, ReSpike achieves end-to-end classification with the following components: 1. an input representation to split spatial and temporal information (Section \ref{subsec:key-res-input-representation}), 2. an ANN-SNN hybrid feature extractor to capture both the spatial and temporal cues from key-residual inputs (Section \ref{subsec:hybrid-feature}), 3. a multi-scale cross-attention fusion module for integrating information from the hybrid branches (Section \ref{subsec:cross-attn-fusion}), and 4. a classification head for final prediction. 

\subsection{Key-Residual Input Representation}\label{subsec:key-res-input-representation}

Given a clip of $T$ frames $\vec{X} = \{\vec{x}[t]\}_{t=1}^T$, we evenly partition them into $n$ segments. The segment length is $s$ (\textit{stride}), hence $T=n\times s$. Each segment contains one key frame for spatial information and $s-1$ residual frames for temporal information. The key frame is the first frame in each segment and the residual frames are obtained by subtracting the current frames from the key frame, given by 
\begin{equation}
    \vec{x}_{res}[t] = \vec{x}[t] - \vec{x}[ks]\text{,  where } k = \lfloor \frac{t}{s} \rfloor. 
\end{equation}
This configuration efficiently addresses motion representation, reducing the need for redundant calculations on similar background regions across the clip. The hyper-parameter stride $s$ provides a trade-off between dense RGB information and the sparse event-like information, \eg, setting $s=T$ will only have one key frame and all the others will become residual frames. 
The choice of $s$ will also affect the energy consumption of the overall ReSpike framework. A higher $s$ will become more energy efficient as it allocates more residuals to SNNs for inference. We conduct an ablation study that shows how the choice of $s$ will affect ReSpike's accuracy and energy efficiency in Section \ref{subsec:ablation}.

\subsection{Hybrid Feature Extractor}\label{subsec:hybrid-feature}

For clarity, the descriptions in the following sub-sections focus on a singular segment of key-residual inputs, though in practice, we process all $n$ segments concurrently in a single forward pass.

In developing the ReSpike architecture, we combine the strengths of both ANN and SNN methodologies for feature extraction, specifically tailored to address the unique characteristics of key-residual representation. 
For each key-residual segment, the key frame $\vec{x}_{key}$ will be fed into the ANN branch; the set of $T^\prime = s-1$ residual frames $\vec{x}_{res}$ will be fed into the SNN branch.

\textbf{ANN branch.} 
Focused on RGB images, this branch extracts global spatial features such as color, pose, shape, and texture.
Since a spatial ANN is essentially an image classiﬁcation architecture, we can build upon the large-scale image classiﬁcation methods. We adopt ResNet-18/50 \cite{he2016deep} initialized with weights pre-trained on ImageNet~\cite{deng2009imagenet} as the backbone. Given a key frame $\vec{x}_{key}$, the global spatial feature $\vec{F}_A \in \reals^{B \times C \times H \times W }$  is extracted from the ANN branch, where $B$ is the batch size and $C \times H \times W $ is the shape of the feature.

\textbf{SNN branch.} 
Targeting event-like residuals, this branch extracts temporal features. Compared to dense RGB images, the inherent sparsity of residuals enables SNNs to capture subtle motion details. Each residual frame is processed in one timestep, the same as processing event-frame data.  Given a set of $T^\prime$ (\ie, $ s-1$) residual frames $\vec{x}_{res}$, the SNN branch processes them sequentially in time order and concatenates the output feature maps across $T^\prime$ total timesteps, resulting in the global temporal feature $\vec{F}_{S} \in \reals^{B \times T^\prime \times D \times H \times W }$, where $D$ denotes the channel dimensionality.

We adopt an SNN backbone based on the MS-ResNet architecture~\cite{hu2024advancing} which mimics residual blocks in the PreAct ResNet~\cite{he2016identity} by building identity mapping from block input to output. 
Notably, among various spiking ResNet architectures
we choose MS-ResNet because it constructs a shortcut through the whole network, maintaining floating point residual transmission (Fig. \ref{fig:network-arch}). This prevents the spike degradation problem commonly found in SNNs~\cite{roy2019towards, kim2021revisiting, hu2024advancing}, preserving information necessary for the later cross-attention calculation. 

\begin{figure}[t]
    \centering
    \includegraphics[width=0.9\textwidth]{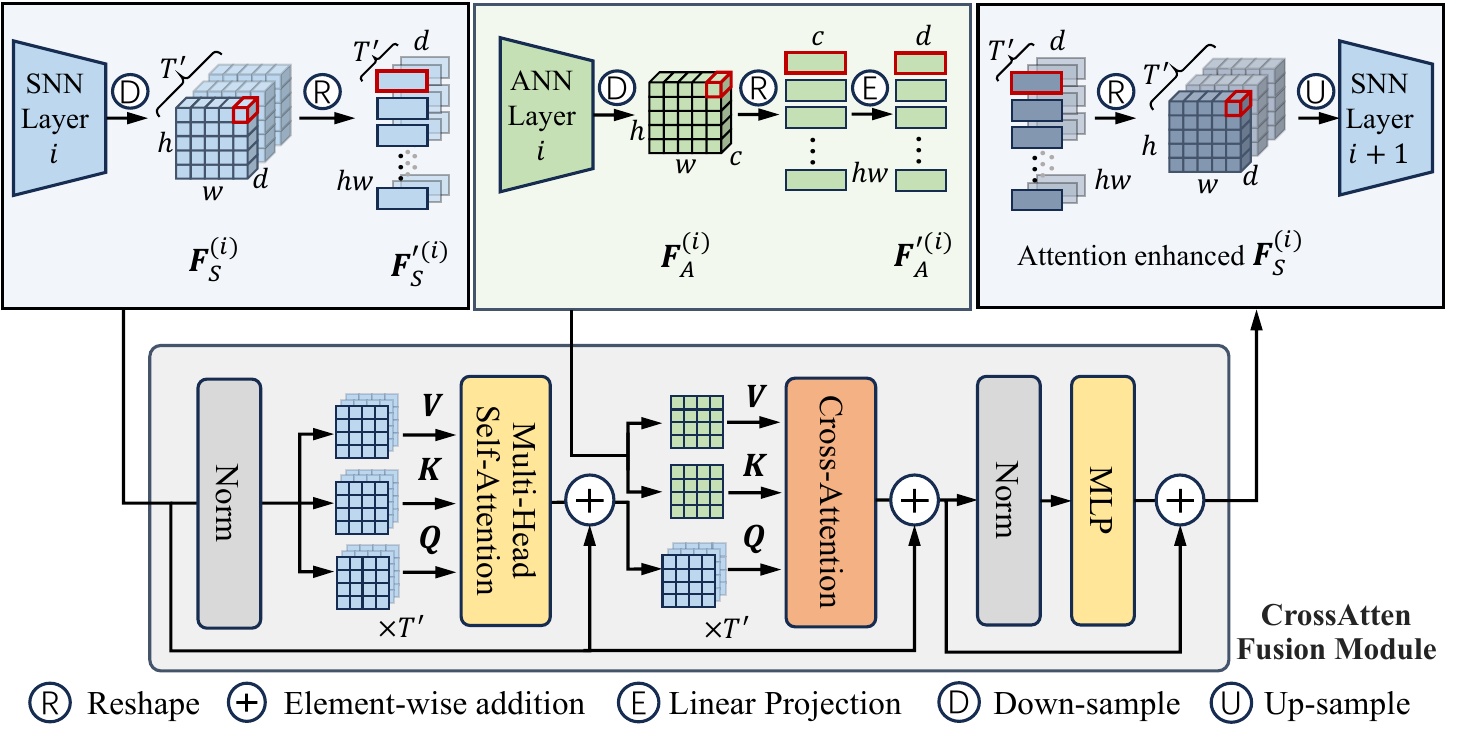}
    \caption{{\bf Cross-attention fusion module} is built upon a standard Transformer block with an additional cross-attention layer, where \textit{Query} map is associated with SNN features and \textit{Key} and \textit{Value} maps are associated with ANN features. Colors indicating the network modules and features are consistent with those in Fig. \ref{fig:network-arch}.}
    \label{fig:transformer}
\end{figure}

\subsection{Spatial-Temporal Feature Fusion} \label{subsec:cross-attn-fusion}

{\bf Multi-scale cross-attention modules.} 
As the SNN branch taking sparse residual inputs lacks awareness of the global spatial information, we can take advantage of the key-frame-speciﬁc inductive biases that the ANN branch offers to enrich the SNN branch with spatial context through inter-modal fusion. We turn SNNs into more ﬂexible spatial-temporal feature extractors by augmenting their underlying MS-ResNet backbone with the cross-attention mechanism, which is effective for incorporating information from other feature modalities \cite{jaegle2021perceiver, jaegle2021perceiverio}. Our key idea is to exploit the self-attention mechanism of Transformers \cite{vaswani2017attention} at multi-scale resolutions 
throughout the SNN branch, building spatial-temporal attention maps at all scales (Fig. \ref{fig:network-arch}). Given the complementary nature of the RGB and residual frames, this fusion module guides the SNN branch to attend to the important semantic features learned by the ANN branch. 

\textbf{Cross-attention layer. }
The cross-attention fusion module is an extension of a traditional Transformer block~\cite{vaswani2017attention, dosovitskiy2010image}, incorporating an additional cross-attention layer. Like a self-attention layer, it is structured around the use of query-key-value (QKV) attention \cite{bahdanau2014neural}. 

As the Transformer architecture takes as input a sequence of tokens, each represented by a vector, we first need to ``tokenize'' the grid-structured feature maps typical of ResNet-based feature extractors. Following the previous studies that applied Transformers to images \cite{rombach2022high, dosovitskiy2010image}, we treat the intermediate feature maps as a set of tokens rather than a spatial grid\footnote{To reduce the complexity of matrix multiplications, we apply a down-sample layer to decrease the feature resolution, hence reducing the token number.}. Given the intermediate ANN feature map $\vec{F}_A^{(i)} \in \reals^{c \times h \times w }$ and SNN feature map $\vec{F}_S^{(i)}[t] \in \reals^{d \times h \times w }$ generated by the ANN and SNN branches at a given layer $i$, we reshape them to be sets of ($h \cdot w$) 2D tokens with latent embedding dimension being the channel dimension. A channel dimensionality reduction through a linear learnable layer is then applied to the ANN tokens. The resulting token inputs 
$\vec{F}_A^{\prime(i)} \in \reals^{(h \cdot w) \times d}$ and $\vec{F}_S^{\prime(i)}[t] \in \reals^{(h \cdot w) \times d}$ are compatible with Transformers' inputs \cite{jaegle2021perceiver}. 

After performing self-attention on the SNN feature  like in a conventional Transformer block we get $\vec{F}_S^{\prime\prime(i)}$, we then calculate the cross-attention 
\begin{equation} \label{eq:attention}
    Attention (\vec{Q}, \vec{K}, \vec{V})=\operatorname{softmax}\left(\frac{\vec{Q} \vec{K}^T}{\sqrt{d}}\right) \vec{V}, 
\end{equation}
\begin{equation} \label{eq:qkv}
    \vec{Q}=\vec{W}_Q^{(i)} \vec{F}_S^{\prime\prime(i)}, \quad \vec{K}=\vec{W}_K^{(i)}  \vec{F}_A^{\prime(i)}, \quad \vec{V}=\vec{W}_V^{(i)} \vec{F}_A^{\prime(i)}.
\end{equation}
Here, $\vec{W}_Q^{(i)} \in \reals^{(h \cdot w) \times d}$  is a learnable weight matrix associated with SNN feature, and  $\vec{W}_K^{(i)} \in \reals^{(h \cdot w) \times d}$ and $\vec{W}_V^{(i)} \in \reals^{(h \cdot w) \times d}$ are weight matrices associated with ANN feature provided as context information. See Fig. \ref{fig:transformer} for a visual depiction.

Since features produced by the SNN branch have an additional time dimension of length $T^\prime$ (\ie, $s-1$), the QKV calculation is applied $T^\prime$ times, each performed between the ANN feature and one slice of the SNN features along the time dimension. In a typical Transformer layer, there are multiple parallel attention `heads', each of which entails generating various $\vec{Q}$, $\vec{K}$, and $\vec{V}$ values per input for Eq. \ref{eq:qkv} and subsequently concatenating the resulting values of $Attention$ from Eq. \ref{eq:attention}. We follow this design for the cross-attention layer implementation. The cross-attention enhanced SNN feature is then reshaped into grid feature maps of dimension $d \times h \times w$ and fed back into the next $i+1$ residual block of the spiking ResNet.

\textbf{Classification Head} 
Given the global spatial feature $\vec{F}_A \in \reals^{B \times C
\times H \times W }$ from the ANN branch and the global temporal feature $\vec{F}_S \in \reals^{B \times T^\prime \times D \times H \times W }$ from the SNN branch, we first take mean the of the temporal feature along the time dimension to get the final output $\vec{F}_S \in \reals^{B  \times D \times H \times W }$, then concatenate it with $\vec{F}_A$ along the channel dimension to obtain fused feature $\vec{F}_F \in \reals^{B \times (C+D) \times H \times W }$. After the fusion module, an Average-Pooling (AP) layer and a Fully-Connected (FC) layer are used for the final prediction of the action class.  The FC layer gives predictions $\hat{\vec{y}} \in \reals^{B \times n_{class}}$. 

\subsection{Training Pipeline} \label{subsec:training-pipeline}
For SNN training, we employ direct coding for spike generation due to its simplicity, adaptability, and proven effectiveness on extensive datasets \cite{kim2022rate}. The training of SNN weights employs the Spatio-Temporal Back-Propagation (STBP) technique \cite{neftci2019surrogate, wu2018spatio}, aggregating gradients across all timesteps. 
During the backward pass, we utilize an arc tangent surrogate gradient function proposed in \cite{fang2021deep} to address the inherent non-differentiability issue of LIF neurons. As such, we can jointly train the SNN branch alongside  the ANN branch and four cross-attention fusion modules in a single forward and backward pass. Cross-entropy loss between the predictions $\hat{\vec{y}}$ and the true labels $\vec{y}$ is applied.

\section{Experiments} \label{sec:experiments}
 {\bf Datasets and metrics.}
We use three widely adopted benchmarks HMDB-51, UCF-101 and Kinetics-400, consisting of 51, 101, and 400 action classes, respectively. HMDB-51 contains 6,766 short clips in total extracted from 3,312 available footages. UCF-101 contains 13,320  clips extracted from 2,500 footages. Kinetics-400~\cite{kay2017kinetics} is a relatively newly introduced large-scale dataset, consisting of over 300k trimmed clips from different sources. We train our models on the training split ($\sim$240k clips) and evaluate on the validation split ($\sim$20k clips).
We train our models from scratch on the three datasets separately, without pretraining on any additional video data. For evaluation, we report the Top-1 mean accuracy over the test subset for HMDB-51/UCF-101 and the validation subset for Kinetics-400. For UCF-101, we report the 3-fold mean accuracy using all 3 splits provided. 

\textbf{Implementation.} We implement the proposed network in PyTorch. The model is trained using a stochastic gradient descent (SGD) optimizer with a momentum of 0.9 and a weight decay factor of 1e-4. The proposed fusion network is trained for 100 epochs with clip length 16 and batch size 32 on an NVIDIA A100 GPU. Code implementation is provided in supplementary materials.

\subsection{Performance Analysis}\label{subsec:sota}
\begin{table}[t]
	\centering
	\caption{Comparing ReSpike against State-of-the-Art ANN-based and SNN-based approaches. }
	\resizebox{0.9\textwidth}{!}{\begin{tabular}{llcccc}
		\toprule
            \textbf{Category} & \textbf{Method} & \textbf{Data} & \textbf{UCF-101} & \textbf{HMDB-51}  & \textbf{Kinetics-400}\\
		\midrule
			\multirow{4}{*}{2D ANN} 
                   & ResNet-18~\cite{he2016deep} & RGB & 76.3 & 48.9 & 52.4 \\
                  & ResNet-50~\cite{he2016deep} & RGB & 80.7 & 55.8 & 61.8 \\
                  & AdaScan~\cite{kar2017adascan} & $\;$RGB$^{\dagger}$ & 78.6 & 41.1  & -  \\   
                  & Residual Two-stream~\cite{christoph2016spatiotemporal} & $\;$RGB$^{\dagger}$  & 82.3 & 43.4 & -  \\
        \midrule
            \multirow{3}{*}{3D ANN} & C3D~\cite{tran2015learning} & RGB & 85.2  & - & 55.6 \\
                  & I3D~\cite{carreira2017quo} & $\;$RGB$^{\dagger}$ & 84.5 & 49.8 & 71.1 \\
                  & Res3D~\cite{tran2017convnet} & RGB & 85.8 & 54.9 & 65.6 \\
        \midrule
            \multirow{5}{*}{SNN} & STS-ResNet~\cite{samadzadeh2023convolutional} & RGB & 42.1 & 21.5  & -\\
                  & SpikeConvFlowNet~\cite{wang2023integrating} & RGB & 40.7 & 23.6 & - \\
                  & MS-ResNet*~\cite{hu2024advancing} & RGB & 51.5 & 21.4 & 34.4 \\
                  & SNN-ResNet*~\cite{li2023uncovering} & RGB & 48.6 & 20.2 & 37.7  \\
                  & TET-ResNet*~\cite{deng2022temporal} & RGB & 54.2 & 23.5 & 36.0\\
        \midrule
            \multirow{3}{*}{\textbf{Ours}} & ReSpike Res-18 ($s=4$)  & RGB (Key-Res) & 77.5 & 52.7 & 61.3  \\
                    & ReSpike Res-50 ($s=4$)  & RGB (Key-Res) & 84.5  & \textbf{58.2} & 69.5 \\
                  & ReSpike Res-50 ($s=2$) & RGB (Key-Res) & \textbf{85.6} & 58.0 & \textbf{70.1} \\
		\bottomrule
	\end{tabular}}
	\label{tbl:sota}
 \vspace{-0.8em}
\end{table}
\begin{table}[t]
    \centering
    \caption{Scaling behavior of ReSpike framework. With stronger backbones, ReSpike performs better. All models take 8 key frames and 8 residual frames with $224 \times 224$ spatial resolution.}
    \resizebox{0.9\textwidth}{!}{\begin{tabular}{llcccc}
    \toprule
    \textbf{Method} & \textbf{Backbone} &  \textbf{HMDB-51}  & \textbf{UCF-101} &\textbf{Kinetics-400} & \textbf{Inference GFLOPs}\\
    \midrule
    ReSpike  ($s=2$) $\quad$ & ResNet-18 & 52.4 & 77.4 & 61.2& 23.8\\
    ReSpike  ($s=2$) & ResNet-50 & 58.0 & 85.6& 70.5 & 47.5\\
    ReSpike  ($s=2$) & ResNet-101 & 59.1 &87.6 & 73.4 & 86.3\\
    ReSpike  ($s=2$) & ResNet-152 & 60.2 &87.9 & 75.1 & 125.0\\
    \bottomrule
    \end{tabular}}
    \label{tab:scaling-law}
    \vspace{-12pt}
\end{table}

\begin{figure}[t]
    \centering
    \includegraphics[width=0.9\textwidth]{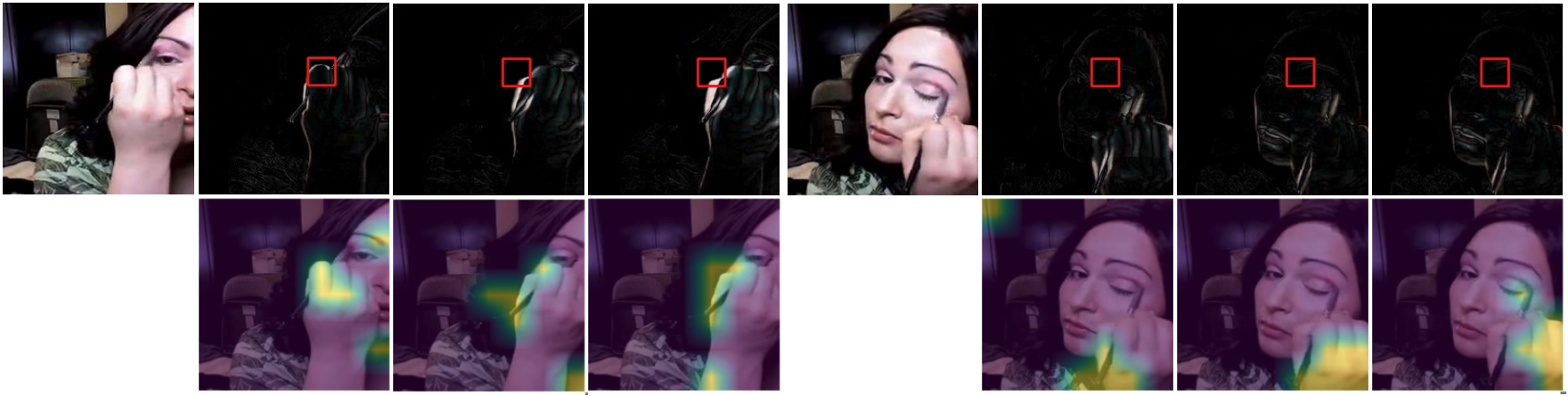}
    \caption{{\bf Attention maps.} This example belongs to action class "Apply Eye Makeup" from UCF-101. For the \textcolor{red}{red} query token associated with the residual frames, we show the top-5 attended key tokens associated with the key frame. ReSpike attends to regions like hand, eye, and makeup brush which are essential for classification.}
    \label{fig:attention-map}
    \vspace{-15pt}
\end{figure}

In this section, we evaluate the effectiveness of our proposed approach against baseline methods, reported in Table \ref{tbl:sota}. We first compare it against ANN-based methods, encompassing two classic categories in literature: (1) 2D ANNs, mianly using standard CNNs as backbone, and (2) 3D ANNs, using 3D convolutional kernels operating on frame volumes to directly generate hierarchical spatio-temporal features.  To ensure a fair comparison with some of them that utilize both RGB clips and optical flow as inputs, we present their results when solely considering RGB inputs, denoted as RGB$^{\dagger}$ in the table. We do not include methods that are pretrained on additional video data. Subsequently, we compare ReSpike with state-of-the-art SNN-based methods~\cite{samadzadeh2023convolutional} and \cite{wang2023integrating}. Considering the scarcity of such approaches in action recognition literature, we also include SNN-based benchmarks tailored for temporal tasks \cite{hu2024advancing, li2023uncovering, deng2022temporal} (* denotes our implementation of these methods). 

\textbf{Kinetics-400.} To the best of our knowledge, we are the first SNN-based method to report on this dataset with a \textbf{70.1}\% accuracy. By utilizing the temporal difference residual frame that aligns with the sparse nature of SNNs, we are able to significantly extend the scalability of directly trained SNNs to a large-scale  dataset. We also acheive comparable performance against prior ANN methods. 

\textbf{UCF-101 and HMDB-51.}
 We observe that ReSpike with the ResNet-50 ANN backbone achieves superior or comparable performance against all ANN methods. Compared to SNN methods, we significantly outperforms them by at least {\bf 31.4\%} on UCF-101 and {\bf 34.6\%} on HMDB-51. 

 \textbf{Scaling behavior.} Our idea of ReSpike is generic, and it can be instantiated with different backbones (\eg \cite{he2016deep, vaswani2017attention}) and implementation specifics. Table \ref{tab:scaling-law} shows ablation on ANN backbones. As backbones continue to evolve, we envision our ReSpike framework closing the gap with leading ANN methods.
 
 \textbf{Attention map visualization.} Our ReSpike architecture (Fig. \ref{fig:network-arch}) consists of 4 cross-attention layers,  each fusing the intermediate SNN and ANN features from the 4 ResNet blocks. Here we present qualitative attention maps from the ﬁnal attention layer (with resolution 7$\times$7) averaged across all attention heads in Fig. \ref{fig:attention-map}. This layer takes in 49 ANN feature tokens and 49 SNN feature tokens as input where each token corresponds to a 32$\times$32 patch in the input modality. We consider a query token that correspond to important motion features (\eg hand/face movement), and then examine the top-5 attention weights associated with the ANN feature. We observe that the semantic regions that correspond to the motion are attended to, providing complementary insights to our model.

\subsection{Model Energy Efficiency} \label{subsec:model-efficiency}

\begin{table}[t]
    \centering
    \caption{Baseline v.s. ReSpike compute energy. Each operation in ANN (FLOP) and SNN (SyOP) consumes $4.6pJ$ and $0.9pJ$, respectively~\cite{horowitz20141}. ``CA'' stands for the cross-attention modules (four in total). All models take 16 frames input with $224 \times 224$ spatial resolution except for TET-ResNet, which takes $112 \times 112$.}
    \resizebox{\textwidth}{!}{\begin{tabular}{lllcccccc}
    \toprule
         \multirow{2}{*}{Category } & \multirow{2}{*}{Architecture} & \multirow{2}{*}{Dataset}  & \multirow{2}{*}{FLOPs ($G$)} & \multirow{2}{*}{SyOPs ($G$)} & \multirow{2}{*}{$\;$ Energy ($mJ$) $\;$} & \multirow{2}{*}{{\bf $\frac{E(.)}{E(ReSpike)}$}}\\
           & & &  &   &  &  \\
    \midrule
          \multirow{2}{*}{2D ANN $\quad$} & ResNet-18~\cite{he2016deep} & - & 29.22 & - & 134.40 & 1.2 \\
          & ResNet-50~\cite{he2016deep} & - & 66.29 & - & 304.92 & 2.8 \\
    \midrule
          \multirow{2}{*}{3D ANN $\;$} &  C3D~\cite{tran2015learning} & - & 154.19	& -	 & 709.27 & 6.4  \\
           & Res3D~\cite{tran2017convnet}	 & - & 163.21  & -	& 750.78 & 6.8 \\
    \midrule
          \multirow{2}{*}{SNN} 
          & MS-ResNet-18*~\cite{hu2024advancing}	& HMDB-51 & 0.35	& 3.19 & 4.47 & 0.04 \\
          & MS-ResNet-18*	& UCF-101 & 0.35	& 3.27 & 4.55 & 0.04 \\
    \midrule
          \multirow{4}{*}{{\bf Ours}} 
          & 	ReSpike Res-18 ($s=4$) & HMDB-51 & 7.25 + 7.22 (CA) & 1.93 & 68.31	&0.62 \\
          & 	ReSpike Res-18 ($s=4$) & UCF-101 & 7.25 + 7.22 (CA) & 1.97 & 68.35	&0.62 \\
        \cmidrule(lr){2-7}
          & {\bf ReSpike Res-50 ($s=4$) \;}& HMDB-51 & {\bf16.35 + 7.22 (CA)}  & {\bf 1.93} & {\bf 110.15} & {\bf 1} \\
          & {\bf ReSpike Res-50 ($s=4$)}& UCF-101 & {\bf16.35 + 7.22 (CA)}  & {\bf 2.01} & {\bf 110.21} & {\bf 1} \\
    \bottomrule
    \end{tabular}}
    \label{tab:energy}
    \vspace{-12pt}
\end{table}

 The energy cost of ReSpike compared to baseline models are reported in Table \ref{tab:energy}, with calculations detailed in Appendix \ref{appendix:energy}.
It is noteworthy that ReSpike's model efficiency significantly surpasses that of ANN-based methods. Compared to our ANN backbone ResNet-18/50, ReSpike achieves 1.2/2.8$\times$ energy reduction. 3D ANNs have energy costs up to 6.4$\times$ and 6.8$\times$ that of ReSpike, underscoring the heavy energy cost of processing frame volumes through 3D convolutions. In contrast, SNNs like the MS-ResNet model display a more favorable energy consumption with a remarkably low ratio of 0.04. By integrating an SNN branch, our hybrid model brings in energy efficiency at higher accuracy.

\subsection{Ablation Studies}\label{subsec:ablation}
We present ablation studies that meticulously dissect the integral components of ReSpike. Unless otherwise mentioned, we set $n=16$ frames per clip with a stride of $s=4$ and thus $T'=3$. 

We vary several design choices. 
Input format variations include: (1) Key: only using key frames (\ie 4 key frames per clip); (2)  Res: only using residual frames (\ie, 12 residual frames per clip); (3) Key + Res: combining key and residual frames as input; and (4) All: using the original 16 RGB frames. 
Network architecture variations include: (1) ANN/SNN only: using only the ANN/SNN ResNet-18/50 as backbone and classifier; and (2)  Hybrid model: using the ReSpike hybrid network with or without the cross-attention module. Note that without the cross-attention module, the ANN and SNN would process data in parallel and fuse only at the classification head. 
In all cases, the network runs a forward pass for each frame (regardless of key frames or residual frames) and conducts average voting across all frames to get the final prediction results. 
We adopt ANN ResNet-18 as the backbone for HMDB-51 and ResNet-50 for UCF-101. 
Results are summarized in Table \ref{tbl:ablation-result} and Fig. \ref{fig:stride-ablation}. 

\begin{minipage}{\textwidth}
  \begin{minipage}[b]{0.63\textwidth}
    \centering
    \captionof{table}{Ablation study on the key-residual data and the hybrid model.}
    \resizebox{\textwidth}{!}{\begin{tabular}{lccccccc}
	\toprule
            \multirow{2}{*}{ID} & \multicolumn{2}{c}{  Architecture }&  \multicolumn{3}{c}{Data} & \multirow{2}{*}{HMDB-51}& \multirow{2}{*}{UCF-101}\\
       \cmidrule(lr){2-3} \cmidrule(lr){4-6} 
               &  Model & Atten &  All &  Key  & Res &  &   \\
		\midrule
		 	E1 & ANN & - & \textcolor{Green}{\checkmark} & \textcolor{red}{$\times$} & \textcolor{red}{$\times$} & 48.86  & 80.70 \\
            E2 & SNN & - & \textcolor{Green}{\checkmark} & \textcolor{red}{$\times$} & \textcolor{red}{$\times$} & 21.35 & 51.54  \\
            E3 & Hybrid & \textcolor{Green}{\checkmark}  & \textcolor{Green}{\checkmark} & \textcolor{red}{$\times$} & \textcolor{red}{$\times$} & 49.45 & 77.31  \\
		\midrule
			E4 & ANN & -& \textcolor{red}{$\times$} &  \textcolor{Green}{\checkmark} & \textcolor{red}{$\times$} & 51.93& 80.52\\
   			E5 & ANN& - &  \textcolor{red}{$\times$} &   \textcolor{red}{$\times$} & \textcolor{Green}{\checkmark} & 34.43  & 44.17 \\
            E6 & SNN & -& \textcolor{red}{$\times$} &  \textcolor{red}{$\times$} & \textcolor{Green}{\checkmark} &  31.88 & 57.52  \\
            E7 & SNN & - & \textcolor{red}{$\times$} &  \textcolor{Green}{\checkmark} & \textcolor{red}{$\times$}& 22.81 & 48.68 \\
		\midrule
            E8 & Hybrid & \textcolor{red}{$\times$} & \textcolor{red}{$\times$} &  \textcolor{Green}{\checkmark} & \textcolor{Green}{\checkmark} & 50.73  & 78.59 \\
            E9 & Hybrid &  \textcolor{Green}{\checkmark} & \textcolor{red}{$\times$} &  \textcolor{Green}{\checkmark} & \textcolor{Green}{\checkmark} & \textbf{52.70}  & \textbf{84.49} \\
		\bottomrule
	\end{tabular}}
	\label{tbl:ablation-result}
    \end{minipage}
    \hfill
    \begin{minipage}[b]{0.36\textwidth}
    \includegraphics[width=\textwidth]{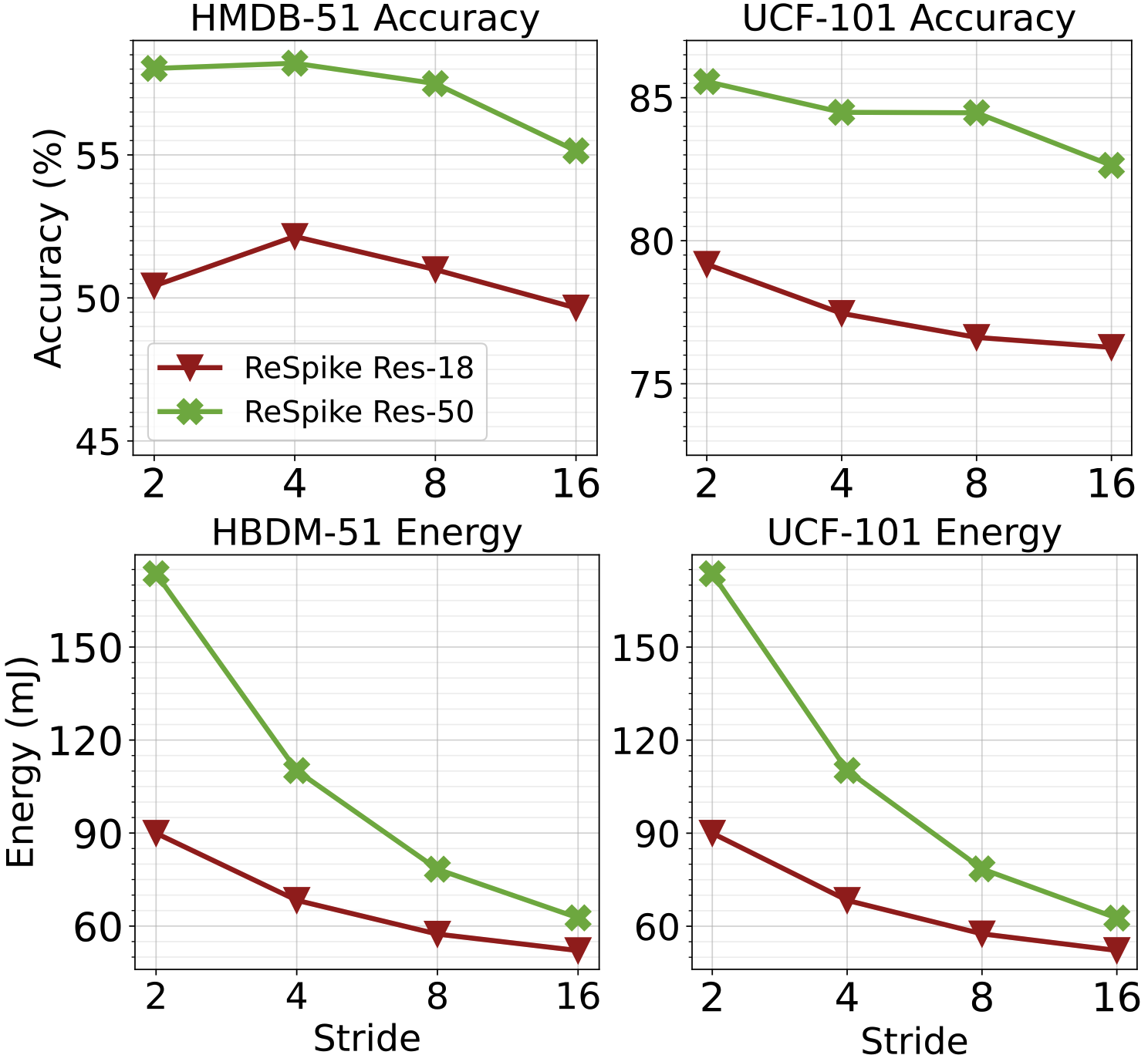}
    \captionof{figure}{Influence of the stride $s$.}
    \label{fig:stride-ablation}
  \end{minipage}
\end{minipage}

\vspace{10pt}
\textbf{Key frame (spatial information).} [E1 vs. E4] shows using key frames only leads to comparable or superior performance compared to using all data, on both HMDB-51 and UCF-101 datasets.
\begin{abox}
    \vspace{-3pt}
   {\small \looseness -1 {\bf Takeaway}: ANNs excel at processing spatial and dense RGB information. Using only key frames sometimes even helps reduce overfitting.}
    \vspace{-3pt}
\end{abox}

\textbf{Residual frame (temporal information). } [E6 vs. E2, E7] SNNs using residual frames only show a significantly better accuracy compared to SNNs using key frames or all RGB frames ($6\sim10$\% improvement on both datasets). In contrast, [E5 vs. E1, E4] shows a large accuracy drop ($15\sim36\%$) when applying ANNs to residual frames. 
\begin{abox}
    \vspace{-3pt}
     {\small{\bf Takeaway}: SNNs are better at handling the sparse event-like information rather than dense RGB information, while ANNs show extremely worse performance in handling residual data.}
    \vspace{-3pt}
\end{abox}

\textbf{Hybrid architecture advantage. } [E9 vs. E4—E7] Hybrid architecture consistently outperforms other single-model single-data configurations. Furthermore, [E9 vs. E1—E3] shows better performance, indicating that key-residual representation is also beneficial. 

\begin{abox}
\vspace{-3pt}
   {\small  {\bf Takeaway}: Hybrid architecture takes advantage of both spatial-temporal information and outperforms all other baseline methods. Moreover, key-residual representation has a synergy with hybrid architecture.}
   \vspace{-12pt}
\end{abox}

\textbf{Cross-attention advantage. } [E9 vs. E8] shows better performances (+5.9\% Acc. on UCF-101) when the cross attention feature fusion module is equipped. 

\begin{abox}
\vspace{-3pt}
    {\small\looseness -1 {\bf Takeaway}: Cross-attention modules selectively focus and fuse relevant information from ANN and SNN branchs, enabling ReSpike to attend to semantic features essential for classification (Fig. \ref{fig:attention-map}).}
\vspace{-3pt}
\end{abox}

\textbf{Influence of the stride variable $s$.} The stride $s$ controls the key and residual partitions. Intuitively, we expect a balanced partition where there's enough RGB information for ANN to learn the spatial semantics, and there are also rich events entailed in residual frames for SNN to learn the temporal dynamics. This is indeed apparent in Fig. \ref{fig:stride-ablation}a, b, where we observe that setting $s=4$ on HMDB-51 gives us the best accuracy. While $s=2$ on UCF-101 shows higher accuracy, it incurs a significantly larger energy cost (Fig. \ref{fig:stride-ablation}d), making $s=4$ the more efficient choice overall.

\section{Conclusion}
In this study, we propose the ReSpike framework that demonstrates the viability of hybrid neural network architectures that leverage both ANNs and SNNs. Through the innovative design of SNN-ANN feature extractors for Key-Residual input and the cross-attention for feature fusion, ReSpike delivers high accuracy and efficiency on challenging datasets, including Kinetics-400. Our findings confirm the potential of combining spatial and temporal processing in a unified framework, opening avenues for future research in high-accuracy and low-energy neuromorphic computing applications for dynamic scene recognition and even more complex vision tasks.


\bibliography{egbib}
\bibliographystyle{unsrt}


\appendix

\section{Energy Calculation} \label{appendix:energy}

We estimate the energy cost of a model based on the number of floating-point operations (FLOPs) executed in 45nm CMOS technology~\cite{rathi2021diet}. 
In ANNs, an operation entails performing a dot-product, which includes one floating-point (FP) multiplication and one FP addition, whereas, in SNN, an operation involves just one FP addition, attributed to the binary nature of spike-based processing. The energy cost for 32-bit ANN operation ($4.6pJ$) is $5.1\times$ more than SNN operation ($0.9pJ$)~\cite{horowitz20141}. 

In ANNs,  the number of FLOPs is given by
\begin{equation}
    \# FLOPs= \begin{cases}k_w \times k_h \times c_{\text {in }} \times h_{\text {out }} \times w_{\text {out }} \times c_{\text {out }}, & \text { Conv layer } \\ f_{\text {in }} \times f_{\text {out }}, & \text { FC layer }\end{cases}
\end{equation}
where $k_w, k_h$ are kernel width and height, $c_{i n}, c_{o u t}$ are the number of input and output channels, $h_{\text {out }}, w_{\text {out }}$ are the height and width of the output feature map, and $f_{\text {in }}, f_{\text {out }}$ are the number of input and output features. 
The number of operations per layer in SNN, denoted as synaptic operations (SyOPs), is given by
\begin{equation}
    \begin{aligned}
& \# SyOPs_l=\text { SpikeRate }_l \times \# FLOPs_l \\
& \text { SpikeRate }_l=\frac{\# \text { TotalSpikes }_l \text { over all inference timesteps }}{\# \text { Neurons }} \\
&
\end{aligned}
\end{equation}
where SpikeRate$_l$ is the total spikes in layer $l$ over all timesteps averaged over the number of neurons in layer $l$. We estimate the SpikeRate on each dataset by averaging across 100 randomly selected samples. For calculating the FLOPs of each layer in the cross-attention modules, we follow \cite{dong2023heatvit}. 

\section{Hardware Memory}

During experiments we found training our SNN-based model requires more memory footprints compared to our baseline 2D ANN-based methods in a single forward pass. With the frame length being set to 16 per clip, ANN ResNet-50 can be trained with a batch size of 256 on an A100 GPU while our method can only be fit into the memory with a batch size of 64 on an A100 GPU and a batch size of 16 on an A5000 GPU. This is primarily due to the additional time dimension of SNN models. SNNs process information over discrete time steps, meaning they maintain and update the state of neurons (e.g., membrane potentials) across all 16 time steps. Training SNNs using methods like Backpropagation Through Time involves unrolling the network over these time steps, necessitating  the storage of intermediate states and gradients for each time step and leading to increased memory consumption.  This is analogous to training recurrent neural networks (RNNs), which also have high memory demands due to the need to retain information across multiple time steps.


\newpage
\section*{NeurIPS Paper Checklist}

\begin{enumerate}

\item {\bf Claims}
    \item[] Question: Do the main claims made in the abstract and introduction accurately reflect the paper's contributions and scope?
    \item[] Answer: \answerYes{} 
    \item[] Justification: In our abstract and introduction, we talk about our contribution on proposing a novel hybrid framework for high-accuracy and low-energy action recognition. We provide extensive experiments on comprehensive datasets to support this claim. We also conduct in-depth ablation studies that verifies the effectiveness of our model design.

\item {\bf Limitations}
    \item[] Question: Does the paper discuss the limitations of the work performed by the authors?
    \item[] Answer: \answerYes{} 
    \item[] Justification: We discuss the limitations in Appendix B, which is about the hardware memory utility.

\item {\bf Theory Assumptions and Proofs}
    \item[] Question: For each theoretical result, does the paper provide the full set of assumptions and a complete (and correct) proof?
    \item[] Answer: \answerNA{} 
    \item[] Justification: Our work does not include theoretical assumptions and proofs.

    \item {\bf Experimental Result Reproducibility}
    \item[] Question: Does the paper fully disclose all the information needed to reproduce the main experimental results of the paper to the extent that it affects the main claims and/or conclusions of the paper (regardless of whether the code and data are provided or not)?
    \item[] Answer: \answerYes{} 
    \item[] Justification: We provide the detailed methodology and experimental setup in Sections \ref{sec:method} and \ref{sec:experiments}. Moreover, we provide all source codes to reproduce the results, including training scripts (detailed configurations included) and evaluation scripts (model checkpoints included). We will open source the code on GitHub after acceptance.

\item {\bf Open access to data and code}
    \item[] Question: Does the paper provide open access to the data and code, with sufficient instructions to faithfully reproduce the main experimental results, as described in supplemental material?
    \item[] Answer: \answerYes{} 
    \item[] Justification: We provide the source codes as supplementary material.

\item {\bf Experimental Setting/Details}
    \item[] Question: Does the paper specify all the training and test details (e.g., data splits, hyperparameters, how they were chosen, type of optimizer, etc.) necessary to understand the results?
    \item[] Answer: \answerYes{} 
    \item[] Justification: Training and test details can be found in Section \ref{sec:experiments}.

\item {\bf Experiment Statistical Significance}
    \item[] Question: Does the paper report error bars suitably and correctly defined or other appropriate information about the statistical significance of the experiments?
    \item[] Answer: \answerNo{} 
    \item[] Justification: We did not include the error bars for 2 reasons. (1) Running experiments on Kinetics-400 with 240k training samples takes a lot of time. (2) We fix the random seed of every experiment, reducing the impact from data loader and other parameters initialization.

\item {\bf Experiments Compute Resources}
    \item[] Question: For each experiment, does the paper provide sufficient information on the computer resources (type of compute workers, memory, time of execution) needed to reproduce the experiments?
    \item[] Answer: \answerYes{} 
    \item[] Justification: We provide the details of computer resources in Section \ref{sec:experiments}. All experiments can be run on a single A100/A5000 GPU. We also provide analysis on inference FLOPs and energy cost in Section \ref{subsec:model-efficiency}.

\item {\bf Code Of Ethics}
    \item[] Question: Does the research conducted in the paper conform, in every respect, with the NeurIPS Code of Ethics \url{https://neurips.cc/public/EthicsGuidelines}?
    \item[] Answer: \answerYes{} 
    \item[] Justification: Our experiments conform to the NeurIPS Code of Ethics.

\item {\bf Broader Impacts}
    \item[] Question: Does the paper discuss both potential positive societal impacts and negative societal impacts of the work performed?
    \item[] Answer: \answerNA{} 
    \item[] Justification: There is no social impact of this work.
    
\item {\bf Safeguards}
    \item[] Question: Does the paper describe safeguards that have been put in place for responsible release of data or models that have a high risk for misuse (e.g., pretrained language models, image generators, or scraped datasets)?
    \item[] Answer: \answerNA{} 
    \item[] Justification: The paper poses no such risks.

\item {\bf Licenses for existing assets}
    \item[] Question: Are the creators or original owners of assets (e.g., code, data, models), used in the paper, properly credited and are the license and terms of use explicitly mentioned and properly respected?
    \item[] Answer: \answerYes{}{} 
    \item[] Justification: Our model and its code development are based on baseline works which are credited in
the paper. Our datasets are the standard benchmarks that are widely used in academia.

\item {\bf New Assets}
    \item[] Question: Are new assets introduced in the paper well documented and is the documentation provided alongside the assets?
    \item[] Answer: \answerNA{} 
    \item[] Justification: The paper does not release new assets.

\item {\bf Crowdsourcing and Research with Human Subjects}
    \item[] Question: For crowdsourcing experiments and research with human subjects, does the paper include the full text of instructions given to participants and screenshots, if applicable, as well as details about compensation (if any)? 
    \item[] Answer: \answerNA{} 
    \item[] Justification: The paper does not involve crowdsourcing nor research with human subjects.

\item {\bf Institutional Review Board (IRB) Approvals or Equivalent for Research with Human Subjects}
    \item[] Question: Does the paper describe potential risks incurred by study participants, whether such risks were disclosed to the subjects, and whether Institutional Review Board (IRB) approvals (or an equivalent approval/review based on the requirements of your country or institution) were obtained?
    \item[] Answer: \answerNA{} 
    \item[] Justification: The paper does not involve crowdsourcing nor research with human subjects.

\end{enumerate}

\end{document}